\pdfoutput=1
\PassOptionsToPackage{square,numbers}{natbib}
\documentclass{article}
\usepackage[preprint]{neurips_2024}
\usepackage[utf8]{inputenc} 
\usepackage[T1]{fontenc}    
\usepackage{url}           
\usepackage{booktabs}   
\usepackage{amsfonts}       
\usepackage{nicefrac}       
\usepackage{microtype}      
\usepackage{xcolor}        
\usepackage{tikz}
\usepackage{forest}
\usetikzlibrary{trees, positioning, shapes, shadows, arrows.meta}
\usepackage{adjustbox}
\usepackage{pgfkeys}
\usepackage{caption} 
\usepackage{geometry} 
\usepackage{amsmath} 
\usepackage{hyperref}
\usepackage{graphicx}
\usepackage[draft,textsize=footnotesize,textwidth=15mm]{todonotes}

\graphicspath{ {./fig/} }
\renewcommand{\footnotesize}{\fontsize{9pt}{11pt}\selectfont}

\makeatletter

\renewcommand{\citep}[1]{\cite{#1}}

\makeatother
\definecolor{darkblue}{rgb}{0, 0, 0.8}
\hypersetup{
	colorlinks=true,
	citecolor=darkblue,
	linkcolor=blue,
	urlcolor=blue
}

\title{{A Comprehensive Survey of Accelerated Generation Techniques in Large Language Models}}

\author{%
    Mahsa Khoshnoodi$^{1}$\quad Vinija Jain$^{2}$\quad Mingye Gao$^{3}$\quad Malavika Srikanth$^{4}$\quad Aman Chadha$^{2,5}$\thanks{Work does not relate to position at Amazon.}\\
    $^1$Amirkabir University of Technology\quad $^2$Stanford University\quad \\
    $^3$Massachusetts Institute of Technology\quad $^4$Columbia University\quad $^5$Amazon GenAI\\
    \texttt{khoshnoodi.ma@gmail.com, hi@vinija.ai, hi@aman.ai}
}

\begin{document}
\setcounter{enumiv}{0} 

\maketitle

\setcounter{footnote}{0} 

\begin{abstract}
    Despite the crucial importance of accelerating text generation in large language models (LLMs) for efficiently producing content, the sequential nature of this process often leads to high inference latency, posing challenges for real-time applications. Various techniques have been proposed and developed to address these challenges and improve efficiency. This paper presents a comprehensive survey of accelerated generation techniques in autoregressive language models, aiming to understand the state-of-the-art methods and their applications. We categorize these techniques into several key areas: speculative decoding, early exiting mechanisms, and non-autoregressive methods. We discuss each category's underlying principles, advantages, limitations, and recent advancements. Through this survey, we aim to offer insights into the current landscape of techniques in LLMs and provide guidance for future research directions in this critical area of natural language processing.\footnote{The relevant papers will be regularly updated at: \url{https://github.com/Arenaa/Accelerated-Generation-Techniques/}}
    
\end{abstract}

    \section{Introduction}\label{sec:intro}
    Inference from large language models (LLMs) requires substantial computational resources owing to several factors. Key among these is the inherent complexity of models like the GPT-family \citep{openai2023gpt4}, LLaMA-family \cite{touvron2023llama}, PaLM \cite{anil2023palm}, OPT \cite{zhang2022opt}, and Mistral \cite{jiang2023mistral} are characterized by their immense complexity, often containing millions or billions of parameters. Consequently, processing input data through the numerous layers of neural networks within these models requires substantial computational resources. Moreover, the inference process is computationally intensive, involving intricate operations like matrix multiplications, nonlinear activations, and attention mechanisms across multiple layers. Additionally, LLMs need large memory allocations due to the extensive data storage within their parameters, including word embeddings and attention matrices.
Furthermore, the autoregressive nature of decoding, where output tokens are generated sequentially based on previously generated ones, limits the potential for parallelization, resulting in slower inference speeds, particularly for longer sequences. Finally, attention mechanisms, commonly employed in LLMs to capture long-range dependencies in input data, contribute to computational complexity, especially when computing attention scores for large input sequences. In conclusion, these factors make inferences from large language models demanding computational resources and time.

Various methods have been developed to address the challenge of speeding up inference from large language models. These techniques includes knowledge distillation \cite{hinton2015distilling, sanh2020distilbert, jiao-etal-2020-tinybert, xu2024survey}, quantization \cite{hubara2016quantized, xu2024onebit, lin2023awq, xiao2024smoothquant}, sparsification \cite{ gale2019state, jaszczur2021sparse, zaheer2021big}, modified attention mechanisms \cite{shazeer2019fast, ainslie2023gqa, kwon2023efficient, ding2023longnet}. However, another critical aspect in enhancing the efficiency of large language models lies in their decoding mechanisms. This survey focuses on these mechanisms within LLMs, exploring and evaluating their role in accelerating inference while maintaining or improving performance.
	
The generation approach in LLMs refers to how these models generate output sequences based on input data. It involves selecting the most likely next token to construct coherent and meaningful sequences at each step. However, accelerating this process poses several challenges. One major challenge is the inherent sequential nature of autoregressive decoding, where each token is generated based on previously generated tokens. This sequential dependency limits the potential for parallelization and leads to slower inference speeds, especially with larger models. Another challenge is maintaining the quality of generated outputs while speeding up the generation process. Any acceleration technique must ensure the generated sequences remain accurate, coherent, and contextually relevant. Speeding up generation should maintain the model's ability to produce high-quality outputs, and the computational resources required can be substantial.
		
In this paper, we comprehensively discuss various accelerated generation techniques. Section \S \ref{sec:speculative} discusses speculative decoding methods, Section \S\ref{sec:early} investigates early exiting methods, and Section \S \ref{sec:non-auto} explores non-autoregressive algorithms (parallel decoding) strategies. Through detailed categorization and in-depth analysis, we provide profound insights into these mechanisms for large language models which highlight strengths, limitations, and future research directions. As illustrated in Figure \ref{fig:taxonomy}, which presents a taxonomy of different algorithms, accelerated generation techniques discussed in this paper are categorized and visualized according to their underlying principles and methodologies.

    \begin{figure}
\hspace{20pt} 

\centering
\tikzset{
    basic/.style = {draw,rounded corners=2pt,  text width=10em, align=center, font=\rmfamily, fill=green!6},
    root/.style  = {basic, rounded corners=2pt, thin, align=center,  rotate=90, child anchor=south, parent anchor=south, anchor=center, text width=14em},
    onode/.style = {basic, thin, rounded corners=2pt, align=center, fill=magenta!5, anchor= east, text width=7em},
    xnode/.style = {basic, thin, rounded corners=2pt, align=center,text width=3cm, fill=magenta!5, anchor= east},
    tnode/.style = {basic, rounded corners=2pt, thin, align=left, text width=14em, fill=blue!7, align=center, font=\small},
    wnode/.style = {basic, thin, rounded corners=2pt, align=center, fill=blue!7, text width=14em, font=\small}
}

\begin{adjustbox}{left=60cm}
\begin{forest}
for tree={
    grow=east,
    reversed=true,
    parent anchor=east,
    base=center,
    ver/.style={basic, rounded corners=2pt, thin, rotate=90, child anchor=north, parent anchor=south, anchor=center},
    edge path={
    \noexpand\path[\forestoption{edge}](!u.parent anchor) -- +(15pt,0) |- (.child anchor)\forestoption{edge label};},
    anchor= east,
	if n children=0{tier=word}{anchor=south}
}
[Accelerated Techniques, ver , l sep=10mm, 
	[Speculative decoding (\S \ref{sec:speculative}), xnode,  l sep=10mm, 
		[Drafting, onode,  l sep=5mm, 
     	[{Blockwise \cite{stern2018blockwise}
           SpecDec \cite{xia2023speculative}
           SpS \cite{chen2023accelerating}
           Speculative Decoding \cite{leviathan2023fast}
           OSD \cite{liu2023online}
           Self-Speculative Decoding \cite{zhang2023draft}
           DistillSpec \cite{zhou2023distillspec}
           REST \cite{he2023rest}
           Cascade Speculative Drafting \cite{chen2023cascade}
           StagedSpec \cite{spector2023accelerating} 
           PaSS \cite{monea2023pass}
           Medusa \cite{cai2024medusa}
           EAGLE \cite{li2024eagle}}, wnode]
]
[Verification, onode, l sep=5mm,
[{SpecInfer \cite{miao2024specinfer}
        SpecTr \cite{sun2024spectr}
        BiLD \cite{kim2023speculative}
        Block-Level Draft Verification \cite{sun2024optimal}
        Multi-Candidate \cite{yang2024multicandidate}}, wnode]
]
[Integrated, onode, l sep=5mm,
      [{Synergy \cite{su2023synergy}
        BiTA \cite{lin2024bita}\\
        SPACE \cite{yi2024generation}
        LLMA \cite{yang2023inference}\\
        Lookahead \cite{zhao2024lookahead}
        SCD \cite{yuan2023speculative}\\
        SARATHI \cite{agrawal2023sarathi}
        SPEED \cite{hooper2024speed}
        TriForce \cite{sun2024triforce}}, wnode]
]
]
[Early Exiting (\S \ref{sec:early}), xnode,  l sep=10mm,
[CALM \cite{schuster2022confident} FREE \cite{bae2023fast}   HASH EE \cite{sun-etal-2022-simple} MPEE \cite{kong-etal-2022-accelerating}  PPD \cite{yang2023predictive}  ConsistentEE \cite{zeng2024consistentee}  SkipDecode \cite{delcorro2023skipdecode} EE-LLM \cite{chen2024eellm}   LayerSkip \cite{elhoushi2024layer}, tnode]
]
[Non-Autoregressive (\S \ref{sec:non-auto}), xnode,  l sep=10mm,
[NAT \cite{gu2018nonautoregressive} ENAT \cite{guo2018nonautoregressive} LT \cite{kaiser2018fast} FlowSeq \cite{ma-etal-2019-flowseq} IR \cite{lee-etal-2018-deterministic} SynST \cite{akoury-etal-2019-syntactically} FCL-NAT \cite{guo2020fine} DePA \cite{zhan2022non} NAG-BERT  \cite{su-etal-2021-non} SAT \cite{wang2018semiautoregressive} NAR-CRF \cite{sun2019fast} Mask-Predict \cite{ghazvininejad2019maskpredict} Parallel Decoding \cite{santilli-etal-2023-accelerating} CLLMs \cite{kou2024cllms} SoT \cite{ning2023skeletonofthought}, tnode]  
]
]\end{forest}
\end{adjustbox}

\caption{Taxonomy of accelerated generation techniques in LLMs is categorized into Speculative Decoding, Early exiting, and Non-Autoregressive methods. Each category presents unique strategies aimed at optimizing model efficiency and efficacy. Speculative decoding explores multiple candidate outputs simultaneously, while early exiting prioritizes termination upon confident predictions. Non-autoregressive methods introduce innovative approaches to parallelization and coherent output generation.}
\end{figure}
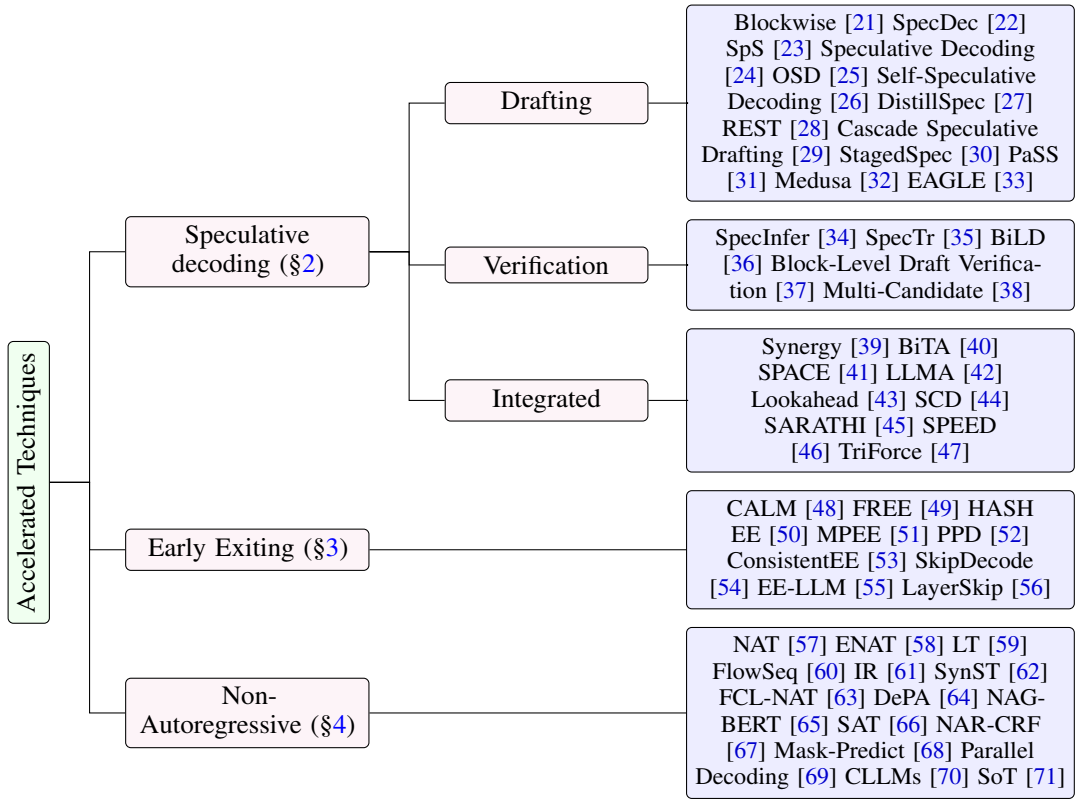
\label{fig:taxonomy}
    
    \section{Speculative Decoding}\label{sec:speculative}
    
Speculative decoding has been inspired by speculative execution  \cite{6312218, hennessy2011computer}, an optimization technique in processors where tasks are executed in parallel to validate their necessity, thereby enhancing concurrency. In the context of text generation by LLMs, speculative decoding is the method that first efficiently predicts multiple tokens and then verifies these predictions simultaneously with the LLMs \cite{miao2023efficient}.

One of the pioneering works in this field, authored by \cite{stern2018blockwise}, introduces a blockwise parallel decoding scheme to enhance the inference process in autoregressive sequence models. Unlike traditional methods that generate outputs sequentially, this approach uses parallel scoring within the model to speed up decoding. For the training part, a multi-output feedforward layer with residual connections was added after the original decoder output layer, and k auxiliary "proposal" models were trained to predict the next k tokens concurrently. These models are executed parallel to predict the next k tokens during inference. The longest prefix of these predictions that would have been generated under greedy decoding is then determined by scoring each position in parallel using the base model. If the length of this prefix exceeds one, one or more iterations of the greedy decoding loop can be skipped. Notably, this approach facilitates faster inference without requiring changes to the model architecture or significant sacrifices in performance.

 \begin{figure}[h!]
    \centering
    \includegraphics[width=0.8\textwidth]{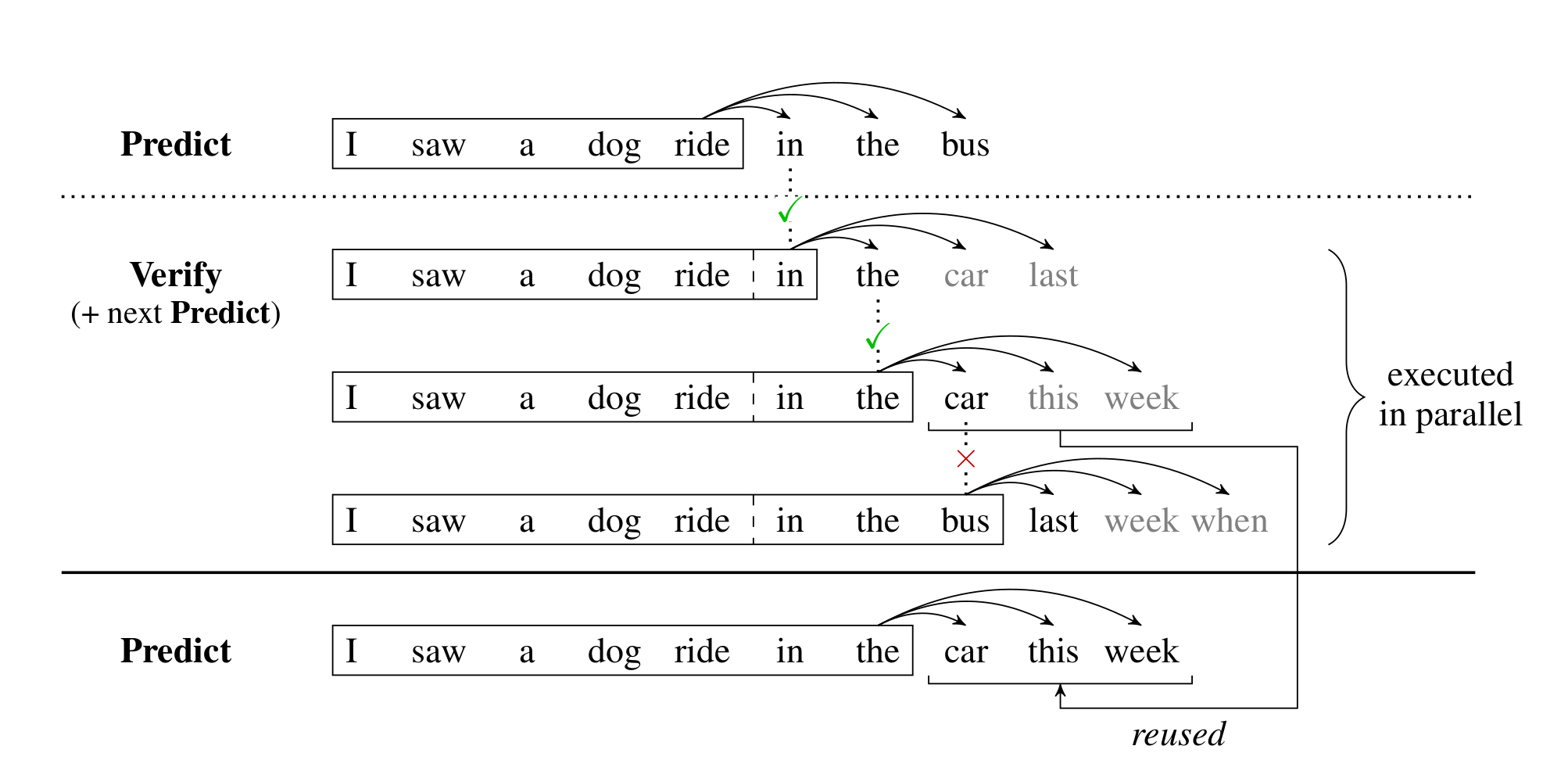}
    \caption{The three core steps of the blockwise algorithm}
    \label{fig:blockwise}
\end{figure}

As illustrated in Figure~\ref{fig:blockwise}, the foundation of this approach is based on three steps: \textbf{predict}, \textbf{verify}, and \textbf{accept}:

\begin{itemize}
    \item In the \textbf{predict} step, the greedy predictions of the primary model and the auxiliary proposal models are generated. As these models are independent, each prediction can be computed concurrently, minimizing the time required compared to a single greedy prediction.
    
    \item During the \textbf{verify} step, the algorithm detects the longest prefix of the proposed \( k \)-length sequence that the base model would have generated.  If the scoring model can process this sequence of \( k \) tokens in fewer than \( k \) steps,  it results in time savings.

	\item Finally, during the \textbf{accept} step, the verified prefix is added to the primary sequence. If the predictions of the base model and the proposal models start to differ, the process stops early to maintain consistency with the output that would have been obtained through greedy decoding.
    
\end{itemize}

In the predict-then-verify paradigm, where multiple future tokens are efficiently drafted and subsequently verified in parallel at each decoding step, \cite{xia2023speculative} introduces SpecDec. This method contains two components—Spec-Drafter and Spec-Verification—that enhance both drafting and verification processes. Spec-Drafter is an independent model trained for efficient and accurate token drafting, focusing on two guiding principles: Capability and Latency. Its objective is to refine the drafting process, improving efficiency without compromising accuracy. Notably, Spec-Drafter demonstrates superior performance to the head-based drafter (as utilized in Blockwise Decoding \cite{stern2018blockwise}) in terms of end-to-end generation quality and efficiency. Spec-verification is a technique that relaxes the verification strategy, accepting tokens that deviate slightly from Auto-regressive (AR) greedy decoding. This flexibility enables the acceptance of a greater number of drafted tokens. Experimental results across various seq2seq tasks, including machine translation and abstractive summarization, underscore the effectiveness of SpecDec. The approach yields approximately a 5x speedup for basic Transformer architecture while maintaining generation quality comparable to beam search decoding. Moreover, SpecDec offers advantages regarding integrating existing models and preserving their original behaviors. These findings underscore the potential of Speculative decoding methods.

In the field of Speculative decoding, two primary phases stand out: drafting and verification. Each phase introduces unique challenges that researchers address through diverse methodologies. In the following section, we explore the approaches used to address them. Subsequently, we investigate methods integrating different approaches to improve Speculative decoding, which we discuss in another section.

\subsection {Methods Targeting Drafting Phase}

The primary challenge in the drafting phase is selecting or training an appropriate draft model. This is because the existing speculative decoding methods require identifying or training a draft model that is well-aligned with the target model. This task becomes particularly challenging when the target model is already fine-tuned for downstream tasks. Meanwhile, running an additional draft model increases the GPU memory overhead, which significantly bottlenecks the deployment on devices with restricted memory capacity.

Leviathan et al. \cite{leviathan2023fast}, and Chen et al. \cite{chen2023accelerating} introduce the idea of training smaller models similarly to the target model as draft models.
In \cite{leviathan2023fast}, authors present speculative decoding, an algorithm to accelerate inference without altering model architecture or training procedures. Using a simpler model to generate multiple tokens, which are then processed in parallel by the target model, maintains the original distribution and improves latency. Results demonstrate a 2X-3X acceleration on T5-XXL compared to T5X, with identical outputs. Concurrently, \cite{chen2023accelerating} addresses the acceleration of LLMs decoding through speculative sampling (SpS). The proposed algorithm accelerates sampling for latency-critical applications by generating multiple tokens from each transformer call. This approach is motivated by the observation that parallel scoring of short continuations, generated by a faster but less powerful draft model, closely matches the latency of sampling a single token from the larger target model. This method involves generating a short draft using a faster auto-regressive model (Draft Model) and scoring it with a larger, more powerful model (Target Model) to replicate the target model's distribution. Experiments conducted using the Chinchilla 70B as the target model and a trained 4B model like Chinchilla as a draft model demonstrate an almost 2.5× speedup.

\textbf{Self-speculative decoding (SSD)} \cite{zhang2023draft} is a novel inference scheme that does not require an auxiliary draft model; instead, it uniquely uses a single LLM for both the drafting and verification stages, thus reducing the total memory usage. In summary, some intermediate layers in the LLM are skipped during the drafting stage, which is selected by Bayesian Optimization; the original LLM is then used to evaluate the drafted token in one forward pass during the verification stage. Although skipping additional layers can accelerate the drafting process in SSD, it also poses the risk of reducing the token acceptance rate during the verification stage, potentially increasing the overall end-to-end inference time. Based on such a consideration, the layer selection process is framed as an optimization problem to minimize the average inference time per token; the inputs to the objective function
 are the number and the specific layers to skip. Since the input space is a power set of layers in the original LLM, the brute force searching method is extremely expensive as numerous layers are included. This problem is solved by employing Bayesian Optimization \cite{Jones1998EfficientGO}. SSD is evaluated on diverse tasks, such as text generation and code generation, with LLaMA 2 and its fine-tuned models.

\textbf{Online Speculative Decoding} (OSD) \cite{liu2023online} introduces a novel approach for online LLM services, which relies on online retraining of draft models. OSD effectively decreases LLM latency by continually updating one or more draft models with user query data observed in real-time. The abundant redundant compute resources in LLM serving clusters, spare FLOPs, are repurposed for online fine-tuning the draft model(s), making the process cost-effective. Existing works of enhancing speculative decoding accuracy typically rely on a fixed draft model after deployment. The main idea of OSD is to continuously retrain (multiple) draft model(s) on observed user query data. As the draft model evolves online, it can adapt to the shifting query distribution, thus significantly enhancing the token acceptance rate. They frame this learning problem based on the aforementioned auxiliary information as online knowledge distillation, where the teacher and student models correspond to the target and draft LLMs in speculative decoding, respectively. They developed a new online learning algorithm based on Generalized Knowledge Distillation (GKD) \cite{agarwal2024onpolicy} to align the draft model with the target model on a newly observed user query. Specifically, the recent queries that the draft model has speculated incorrectly are continuously tracked, and the draft model is forced to emulate the target model’s outputs on these queries. 

In OSD, the draft model qq is first trained on the warmup dataset DD. For each request the serving cluster receives, the pre-trained draft model $q_\theta$ speculates multiple tokens until getting <EOS> token. Once a generated token is rejected by the target model PP, both the draft and target distribution are stored in a temporal buffer RR. A reply buffer QQ tracks the RR for each request. For every II iteration, the draft model is updated on QQ to minimize the distillation loss ll, which is determined by the batch input and the specific model pairs (p, $q_\theta$). OSD is evaluated on both synthetic and real query data using several popular LLMs. It demonstrates that OSD yields a quickly increasing token acceptance rate as the draft model is exposed to more query data. Compared to the offline setting, OSD substantially improves the token acceptance rate by 0.1 to 0.65, translating into 1.22× to 3.06× latency reduction. Experiments also demonstrate that OSD’s acceptance rate first drops when the query distribution changes but swiftly rebounds as more data is processed, implying OSD’s strong adaptability to shifting input distributions.

Concurrently, \textbf{DistillSpec} \cite{zhou2023distillspec} also addresses the challenge of aligning a smaller draft model with a larger target model with knowledge distillation (KD) \cite{hinton2015distilling}.  Existing works of the KD primarily focus on enhancing the performance of the small student model; DistillSpec, in contrast, aims at improving the acceptance rate during Speculative Decoding by aligning the student model with the teacher model, which is the draft model and the target model in SD, respectively. Specifically, DistillSpec uniquely uses the draft model’s on-policy data in the distillation process and tailors the divergence function to the task and decoding strategy, which are proven to be crucial for improving the draft model alignment with the target model. 

Let the draft model $M_q$ be parameterized by $\theta$. DistillSpec utilizes predictions from the target model $M_p$ as a source of teacher while training $M_q$ (student). Given a divergence function $D$ that measures the misalignment between two distributions, the training of $M_q$ based on KL divergence seeks to minimize the token-level distribution of $M_p(p(y_t))$ and $M_q(q(y_t))$ over a training set $G$. Choices for $G$ and $D$ are guided by how the distilled draft model improves the SD speed. Maximizing the sequence-level acceptance rate $\alpha(x)$ is equivalent to minimizing the expected total variation distance $(D_{TVD})$ between $p(y_t)$ and $q(y_t)$ over the output sequence distribution of $M_p$. DistillSpec shows substantial speedups of 10-45\% over standard SD methods across various benchmarks using different sampling strategies. It also demonstrates the ability to reduce decoding latency by 6-10 times with minimal performance drop, particularly effective in practical scenarios with models of varying sizes.

In \textbf{REST} \cite{he2023rest}, unlike traditional speculative decoding methods that use a smaller language model for draft token generation, it employs a non-parametric retrieval datastore, allowing easy integration with any LLM and accelerating inference without additional training. REST uses previous tokens as queries to identify matches in a datastore, selecting subsequent tokens as candidates. A Trie structure is used to select high-frequency prefixes as draft tokens, which are then verified by the LLM through a single forward pass using a tree attention mask. Extensive experiments show significant speed improvements. For instance, REST achieved a 2.12$\times$ to 2.36$\times$ speedup on HumanEval with CodeLlama models and a 1.62$\times$ to 1.77$\times$ speedup on MT-Bench \cite{zheng2023judging}  with Vicuna models \cite{vicuna2023}. These results highlight REST's effectiveness in various domains and its potential to accelerate LLM generation processes.

In \cite{chen2023cascade}, the paper introduces a new algorithm, \textbf{Cascade Speculative Drafting (CS Drafting)}, addressing the primary issue of slow autoregressive generation in the drafting phase of speculative decoding. The authors propose the \textbf{Vertical Cascade} approach to overcome this challenge and improve LLMs inference. A significant inefficiency in speculative decoding arises from relying on autoregressive generation by a smaller draft model. Since the draft model runs $k$ times for each target model run, the cost remains considerable despite its smaller size. To address this inefficiency, they incorporate an even smaller model to assist in drafting, whereas the original draft model reviews the generations of this smaller model. This process is repeated recursively until it reaches a statistical draft model, such as a Bigram language model, with minimal computational costs. Each recursion step aims to reduce the drafting latency without altering the output distribution.

Furthermore, not all drafting tokens are equally important during the drafting steps of speculative decoding. Motivated by this observation,  the \textbf{Horizontal Cascade} is introduced, which improves the time allocation by adjusting the draft token allocation. Here, the largest draft model is assigned to generate the first draft token because it holds the highest likelihood of acceptance. Subsequently, smaller draft models were employed as the likelihood of token acceptance decreased. This process concludes with the smallest model, a statistical language model, further reducing the time cost of generating less critical draft tokens with a resource-intensive draft model and ultimately minimizing the overall latency. Experiments show that this new method achieves up to 81 percent additional speedup over speculative decoding while maintaining the same output distribution as the target model. This work demonstrates that LLM inference can be further sped up by cascades without sacrificing the generation quality or requiring additional task-specific training.

\textbf{PaSS} \cite{monea2023pass} introduces Parallel Speculative Sampling (PaSS), a novel method aimed at speeding up decoding by generating multiple tokens simultaneously. Traditional speculative sampling methods typically use two separate models: a large target model and a smaller, quicker draft model. However, this setup is inefficient in terms of memory and computational resources. PaSS addresses this inefficiency by employing parallel decoding from a single target model to draft multiple tokens concurrently. It incorporates \textbf{"look-ahead"} embeddings into the model architecture to enable parallel generation. Importantly, only the look-ahead embeddings are trained, while the target model remains fixed. PaSS utilizes rejection sampling techniques to ensure that the generated tokens match the target distribution, guaranteeing at least one new token per iteration. Experimental results demonstrate that PaSS achieves up to a 30\% improvement in decoding speed compared to the autoregressive decoding baseline, all without any degradation in quality. Additionally, PaSS exhibits clear advantages over using a fixed "[UNK]" look-ahead token, further emphasizing its effectiveness in accelerating decoding from large language models.
		
Spector et al. introduce \textbf{Staged Speculative Decoding} \cite{spector2023accelerating}, an algorithm designed to accelerate inference for LLMs in on-device, small-batch scenarios. The primary aim is to improve latency, enable personalization, and enhance privacy by executing LLMs locally rather than relying on cloud-based solutions. This approach innovatively enhances speculative decoding by increasing batch size/quality and speeding up the draft model. Unique techniques include introducing tree-structured batches and two-stage speculation, previously unexplored in the field. The algorithm consists of two main components: restructuring the speculative batch as a tree to reduce generation costs and increase tokens per batch and incorporating a second stage of speculative decoding by involving the draft model. Experimental results using a 762M parameter GPT-2-Large model \cite{radford2019language} demonstrate the effectiveness of staged speculative decoding, achieving a 3x reduction in bandwidth usage and a 3.16x performance improvement compared to standard greedy decoding while maintaining output quality. Speedups vary based on prompt complexity, ranging from 2-10x, highlighting significant advancements in enhancing the efficiency and applicability of LLMs in on-device scenarios.

\textbf{Medusa} \cite{cai2024medusa} address the challenge of finding the ideal draft model by adding additional decoding heads to the base model,  inspired by Blockwise Decoding \cite{stern2018blockwise}. The idea's unique value proposition lies in training multiple decoding heads, termed "Medusa heads," on the same model instead of introducing a new draft model. Medusa heads can be trained concurrently with the original model, which remains frozen during training and generates blocks of tokens at each decoding step. This allows for fine-tuning large models on a single GPU, using the powerful base model's learned representations. Since the heads are a single layer akin to the original language model head, Medusa does not add complexity to the serving system design and is suitable for distributed settings. Moreover, instead of the traditional importance sampling scheme, the authors propose an efficient and high-quality alternative by relaxing the requirement of matching the distribution of the original model. This method becomes more efficient at higher sampling temperatures, allowing for longer accepted sequences.

Medusa introduces multiple heads on top of the last hidden states of the LLM, enabling the prediction of several subsequent tokens in parallel. Each Medusa head is a single-layer feed-forward network augmented with a residual connection. Training these heads utilizes the same corpus that trained the original model or a new corpus generated using the model itself. Importantly, the original model remains static during this training phase, with only the Medusa heads fine-tuned. For Medusa heads, the top-1 accuracy for predicting the 'next-next' token hovers around 60\%, while the top-5 accuracy exceeds 80\%. This substantial increase indicates the potential to significantly amplify the number of tokens generated per decoding step by strategically leveraging the multiple top-ranked predictions made by the Medusa heads. During inference, candidates are crafted by taking the Cartesian product of multiple top predictions from each Medusa head, and these candidates are processed in parallel using a tree-based attention mechanism.

The authors introduce the typical acceptance scheme to select plausible candidates, employing a combination of hard and entropy-dependent thresholds similar to truncation sampling. They always accept the first token using greedy decoding, ensuring that at least one token is generated in each step. The final output is then the longest sequence that passes the acceptance test. Medusa tested with Vicuna models—specialized LLaMA models fine-tuned for chat applications—with parameter counts of 7B, 13B, and 33B. To train Medusa heads, they utilized the publicly available ShareGPT dataset, a subset of the training data originally used for Vicuna models, and trained for a single epoch. Evaluation using the MT-bench consistently showed that Medusa achieved approximately a 2x speedup in wall time across a broad spectrum of use cases. With Medusa's optimization, a 33B parameter Vicuna model could operate as swiftly as a 13B model.

\textbf{EAGLE} \cite{li2024eagle} introduces a new drafting approach named "Extrapolation Algorithm for Greater Language-model Efficiency". This algorithm is devised based on two crucial observations stemming from reevaluating speculative sampling techniques. Firstly, it recognizes that autoregression at the feature level proves to be simpler than at the token level. In contrast to token sequences, which represent straightforward conversions of natural language, feature sequences demonstrate greater consistency. Employing autoregressive processing at the feature level and subsequently generating tokens using the language model head of the original LLM produces superior outcomes compared to directly predicting tokens autoregressively. Secondly, it acknowledges that the inherent uncertainty in the sampling process significantly limits the predictability of the next feature. EAGLE operates by conducting the drafting process autoregressively at the more structured (second-to-top-layer) feature level. It addresses sampling uncertainty in predicting the next feature by incorporating tokens from one time step ahead. Remarkably, EAGLE ensures the preservation of the output distribution of the LLM while markedly enhancing generation speed.

\textbf{Decoding Speculative Decoding} \cite{yan2024decoding}  aims to investigate the most effective method for selecting and designing draft models to improve speculative decoding efficiency. Various factors influencing speculative decoding and their impact on speedups are analyzed. Subsequently, an analytical model is proposed as a framework to quantify the throughput of a speculative decoding model, aiding in selecting the appropriate draft model for a specific workload. Among these factors, the Token Acceptance Rate (TAR) is a crucial metric. TAR delineates the average number of tokens generated by the draft model that are accepted by the target Large Language Model (LLM).
In the experiments, LLaMA-65B and OPT-66B, along with wider variants of the LLaMA and OPT families, are utilized as target and draft models, respectively. The draft models vary in size, with parameters ranging from approximately 5× to 528× fewer than the target models. Results reveal that smaller models with lower Total Arithmetic Requirement (TAR) yield higher throughput, suggesting that higher TAR does not consistently result in the highest throughput. Subsequently, the paper introduces an analytical model designed to predict the throughput of speculative decoding. This model elucidates the interplay between different factors such as model latency and TAR and their impact on speculative decoding throughput. Furthermore, they demonstrate how users can use their analytical model to make well-informed choices regarding draft model selection.

\subsection{Methods Targeting Verification Phase}

One of the primary aspects of Speculative Decoding is to ensure that the distribution of the target model is preserved while generating multiple tokens with a faster, less powerful draft model. Therefore, the verification criterion is crucial for determining the validity of drafted tokens in the generated sequence. Given an input sequence $x_1, \ldots, x_t$, a set of drafted tokens $x^\prime_{1} \ldots, x^\prime_{K}$, and the corresponding probability distributions $p_1, \ldots, p_K$ (target) and $q_1, \ldots, q_K$ (draft), the verification criterion for the $i^{th}$ drafted token can be expressed as follows:
\[ x^\prime_{i} = \text{arg max} \, p_i, \]

where $i = 1, \ldots, K$, This criterion selects the token with the highest probability according to the computed distribution $q_i$, which is based on greedy decoding. Although greedy decoding is simple and clear, the strict matching requirement of this criterion often results in the rejection of high-quality drafted tokens simply because they differ from the top-1 predictions of the target LLM, thereby constraining the speedup of the paradigm. Therefore, some approaches use top-k sampling \cite{fan2018hierarchical} to relax greedy matching criteria\cite{stern2018blockwise, xia2023speculative}. Top-k sampling randomly selects from the k most likely candidates for the next word in a sequence, providing diversity and reducing repetition compared to greedy decoding.

\textbf{Speculative sampling} introduced by \cite{leviathan2023fast, chen2023accelerating}, is a novel method designed to ensure that the target model distribution is preserved. This method works by sampling from a simpler and more efficient model distribution $q(x)$ instead of directly sampling from the target distribution $p(x)$. The sample is maintained if the sampled value from $q(x)$ is less than or equal to the corresponding value from $p(x)$. However, if the sampled value from $q(x)$ is greater than the corresponding value from $p(x)$, the sample is rejected with a probability of $1 - \frac{p(x)}{q(x)}$. In such cases, the sample is resampled from an adjusted distribution $p'(x) = \text{norm}(\max(0, p(x) - q(x)))$, which is computed as the difference between $p(x)$ and $q(x)$ and normalized to ensure a valid probability distribution. This approach ensures that between 1 and $1+N$ tokens are generated per run, preventing an increase in the number of sequence runs compared to conventional approaches. This method theoretically demonstrates that it can maintain output distributions identical to the target LLM.

\textbf{SpecInfer} \cite{miao2024specinfer} presents a system that enhances LLMs' end-to-end latency and computational efficiency by employing tree-based speculative inference and verification. Instead of focusing on a single sequence of tokens, SpecInfer generates and verifies a token tree, where each node represents a unique token sequence. Previous approaches typically relied on a single small-language model for speculation, which may not be well-aligned with a target LLM because of the capacity gap between them. By contrast, SpecInfer maximizes speculative performance by simultaneously considering a variety of token sequences organized in a tree structure for a given input prompt. The system introduces an expansion-based and merge-based mechanism for constructing token trees by exploiting the diversity within a single small speculative model (SSM) and across multiple SSMs.

Furthermore, it introduces a token-tree verifier that accepts a token tree generated by the speculator as input and verifies the accuracy of the tokens against the output of a target model. One of the challenges that must be addressed in token tree verification is computing the attention scores for all sequences of the token tree efficiently. To address this challenge, the authors introduced tree attention, which generalizes the attention mechanism from sequences to tree structures. In addition, SpecInfer develops a tree-based parallel decoding mechanism for computing tree attention for all tokens in a token tree in parallel. For a given speculated token tree $\mathcal{N}$, it uses the tree-based parallel decoding to compute its tree attention and generates an output tensor $\mathcal{O}$ that includes a token for each node $u \in \mathcal{N}$. Experiments show that SpecInfer outperforms existing LLMs by 1.5-2.8× for distributed inference and 2.6-3.5× for offloading-based inference while maintaining the same generative performance.

The \textbf{Big Little Decoder (BiLD)} method, as presented in \cite{kim2023speculative}, proposes another approach to aligning the predictions of the small model with those of the large model. The BiLD framework comprises a small and a large model, along with a policy for the verification phase that determines which model to utilize for each decoding iteration. This policy consists of two components: fallback and rollback. The primary principle of the policy is that the small model should have the ability to decide when to transfer control to the large model. When the small model lacks confidence in its prediction, it is preferable to let the large model take over. If the small model exhibits over-confidence in its incorrect predictions, this can have repercussions since a single incorrect prediction can impact all subsequent token predictions. To prevent such scenarios, it is advantageous to have the large model scrutinize the predictions of the small model and validate each one. The rollback policy is employed to revert the predictions of the small model if they diverge from those of the large model.

\textbf{SpecTr} \cite{sun2024spectr} extends previous work on speculative decoding by using optimal transport theory to further speed up the decoding process. In speculative decoding, the draft selection phase has two main goals. Firstly, it ensures that the chosen token aligns with the probabilities of the large model, maintaining the quality of the decoded output without any degradation compared to that of the large model. Secondly, it aims to maximize acceptance, meaning that the higher the likelihood we accept a token, the more we can speed up processing through parallelization. To achieve these goals, SpecTr formulates the draft selection problem as a discrete optimal transport problem with membership costs (OTM), enabling the simultaneous consideration of multiple draft sequences. This approach maximizes the likelihood of selecting validated tokens. Additionally, OTM extends the token-level selection to the sequence level, facilitating the concurrent optimization of multiple drafts. While the optimal plan can be computed via linear programming, SpecTr introduces an efficient approximate algorithm called K-SEQ, which can be computed almost linearly in time. Experimental evaluations conducted with state-of-the-art transformer models demonstrate the efficacy of SpecTr. It achieves a notable 2.13x speedup over autoregressive decoding and a 1.37x improvement over speculative decoding, all without compromising output quality.

Inspired by SpecTr, a new perspective for the verification phase is introduced by \textbf{Optimal Block-Level Draft Verification} \cite{sun2024optimal}. A unique formulation of the verification problem within speculative decoding is presented, framing it as a block-level optimal transport (OT) problem with a specific cost function. This formulation aims to enhance the expected token efficiency within a draft block directly. Two challenges are encountered by the proposed verification algorithm. During speculative decoding, some constraints remain unknown due to the causal structure of language modeling, necessitating additional calls to both small- and large-language models for naive solutions. Additionally, even if these constraints are known, solving them using OT solvers could require exponentially proportional time to the block length, which is often impractical. These challenges are overcome by introducing a new computationally efficient algorithm that optimally solves the block-level transport problem. Notably, the introduced verification algorithm does not require any extra calls to draft large models. In the experiments conducted, it is empirically compared to standard speculative decoding methods. Across various tasks and datasets, consistent improvements over the previous speculative decoding approaches are observed.

\textbf{Multi-Candidate Speculative Decoding} \cite{yang2024multicandidate} suggests a strategy of sampling multiple candidates from a draft model and organizing them into batches for verification. While this approach seems straightforward, it encounters the challenge of speculative decoding, which cannot directly leverage multiple candidates to enhance acceptance rates without compromising the output distribution of the target model. To tackle this issue, they propose a multi-candidate verification algorithm. The backbone of LLMs, the transformers, generates text via causal language modeling. It uses the keys and values of previous tokens to generate new ones. The fact that these keys and values must be sent to the computational unit at each model increases the overhead for usage in multiple candidate verification processes. To mitigate this, the authors use Tree attention, which enables multiple candidates to share the caches of generated tokens. Additionally, since multiple candidates sampled from the draft model may collide, they also introduce a more efficient version for candidates sampled without replacement. In fact, a portion of potential computing resources have not been fully utilized. This approach involves utilizing this portion of resources to perform parallel verification on another dimension (i.e., the batch dimension) to improve the acceptance rate of draft tokens significantly.

\subsection{Integrated Methods for Enhanced Speculative Decoding}

Several studies combine speculative decoding with other methods to enhance its efficiency. This integration of speculative decoding with supplementary techniques is pursued to overcome inherent complexities and boost the effectiveness of this approach. In the following section, we analyze these studies, exploring the creative methods and achievements of blending speculative decoding with various approaches.

 \cite{su2023synergy} explores the combined impact of speculative decoding and batching on accelerating inference for serving LLMs like GPT-3. Motivated by the low parallelism and hardware utilization observed in LLM serving, the study aims to investigate how speculative decoding and batching can address these challenges and enhance performance. The interaction between these two techniques has not been thoroughly understood previously. To address this, the authors conduct a comprehensive profiling study and modeling analysis of speculative decoding under different batch sizes. They propose an adaptive method to select the speculation length based on the batch size. The proposed approach is implemented in a prototype system using OPT and LLaMA transformers models. The results highlight that the optimal speculation length varies depending on the batch size, and the adaptive method outperforms fixed speculation lengths. In dynamic traffic scenarios, the adaptive method achieves up to a 9\% reduction in latency compared to fixed schemes. This study sheds light on the potential synergy between speculative decoding and batching, offering insights into optimizing inference processes for serving large language models efficiently.
 
\textbf{BiTA} \cite{lin2024bita} utilizes the concept of prompt tuning to achieve lossless Semi-autoregressive (SAR) decoding for Autoregressive (AR) language models with minimal additional learnable parameters. This method allows a transformer-based AR model to adopt a SAR generation style through efficient tuning. BiTA consists of two main components: SAR draft generation via bidirectional tuning and streamlined verification of the generated draft candidates. Bi-directional tuning incorporates prompt and mask tokens to predict future tokens, extending beyond the next token for AR models. Metaphorically, this approach employs learnable prefixes and suffix embeddings in token sequences. An elaborate tree-based attention mechanism enables simultaneous generation and verification in a single forward pass within the converted model, eliminating the need for additional validation steps or external verification models. By leveraging prompt tuning, the proposed method in BiTA serves as a plug-and-play module for accelerating publicly available transformer-based LLMs, particularly well-trained chatbots, without compromising their strong generative capabilities.

\textbf{SPACE} \cite{yi2024generation} also adopts a similar strategy by semi-autoregressive supervised fine-tuning (SAR-SFT) the target model to predict future token sequences in parallel, eliminating the dependency on extra small models. SAR-SFT aims to enable the LLM to make educated guesses about upcoming tokens rather than accurately predicting multiple tokens in parallel. This strategy aims to improve the model's inference efficiency by preparing likely token sequences in advance, which can later be validated and refined by the auto-correct decoding algorithm. Since SPACE uses the same LLM for both generating and verifying candidate tokens, this algorithm has been developed to enable the unified LLM to concurrently verify tokens from the current step and generate new candidates for the next step within a single forward pass. Experiments conducted across various LLMs with parameters ranging from 6B to 70B validate the efficacy of SPACE. It achieves an impressive inference speedup ranging from 2.7x to 4.0x in HumanEval-X while maintaining output quality. This underscores the significance of SPACE in addressing the challenges of serving LLMs efficiently in real-world edge server environments.

\textbf{LLMA} \cite{yang2023inference} introduces a method that utilizes a prefix-matching approach to retrieve content from a document database. LLMA leverages the overlap between the LLM's output and a reference, which is common in various practical scenarios. Therefore, LLMA does not require additional models for the token drafting process. Applications such as retrieval-augmented generation and multi-turn conversations often have substantial text overlaps that LLMA utilizes during decoding. Initially, LLMA chooses a text span from the reference, copies its tokens to the LLM decoder, and then evaluates them based on the probabilities of the output tokens, which can be efficiently done in parallel. Experiments conducted on retrieval-augmented generation and cache-assisted generation demonstrated that LLMA achieved a 2x speedup on 7 B, 13 B, and 30 B LLaMA models without sacrificing accuracy. The impact of different copy lengths was analyzed, showing that longer lengths enhance parallelism and reduce decoding steps, with optimal performance reaching a length of 15. The method demonstrates robust performance across various model sizes and application scenarios where text overlaps exist between the inputs and outputs.

\textbf{Lookahead} \cite{zhao2024lookahead} extends the idea of using a RAG system instead of a draft model. The authors introduce a multi-branch strategy that retrieves multiple
drafts simultaneously. These drafts are then efficiently decoded
and validated in parallel through the Verification and Accept (VA)
process. The VA process then identifies the correct
sub-sequence for each draft and retains the longest sub-sequence
as the output tokens. The trie-based Retrieval (TR) process enables the simultaneous generation of multiple branches, each representing a sequence of tokens. Subsequently, a verification and Accept (VA) process was performed for each branch to identify the longest correct sub-sequence as the final output. A trie tree is a data structure that efficiently handles prefix matching by organizing nodes as individual characters or words. In this study, each tree node represents a token ID and a path from the root to a leaf node represents a branch.

Before and after each step, which includes the draft retrieval process and the VA process, a global trie tree will be updated through multiple procedures such as "Branch Inserting", "Branch Eliminating", and "Node pruning". During the retrieval process, the trie tree is queried to provide drafts. The Trie Tree Retrieval method involves extracting multiple branches from a trie tree using a prefix, which is a sequence of tokens. The length of the prefix determines the quantity and relevance of the retrieved branches: shorter prefixes yield more branches, while longer ones result in more closely related branches. A multi-stage retrieval strategy is employed to balance branch count and correlation. Initially, attempts are made to match longer prefixes, adjusting the prefix length if the number of tokens associated with matched branches is significantly smaller than desired. If the count of matched branches falls below a threshold, all are used for the Verification and Acceptance (VA) process; otherwise, tokens with the highest frequency are chosen. Branches originating from the input prompt are prioritized by amplifying their frequency. Implementing Lookahead with a parallel multi-branch draft further enhances AntGLM-10B’s average inference speed to 263.4 tokens/s, achieving a remarkable 5.03 times speed-up, which represents a significant acceleration.

\textbf{Speculative Contrastive Decoding (SCD)} \cite{yuan2023speculative} combines the idea of speculative decoding with contrastive decoding, leveraging the inherent contrast between amateur and expert models in speculative decoding. In its implementation, SCD utilizes an amateur model to generate multiple tokens, which are then reviewed by an expert model akin to speculative decoding. However, unlike traditional speculative decoding, SCD employs a contrasted distribution based on both the amateur and expert models during the token review. Tokens are probabilistically rejected and resampled based on this contrasted distribution. Experimental results conducted on four benchmarks using LLaMA 2 models demonstrate that SCD achieves similar acceleration as speculative decoding (1.5-3x) while enhancing metrics such as perplexity, accuracy, and human evaluation scores compared to the expert LLaMA 2 model. Further analysis indicates that SCD accepts easier tokens and rejects harder ones, resulting in greater benefit from contrast.

\textbf{SARATHI} \cite{agrawal2023sarathi} proposes a new method by employing chunked-prefills and decode-maximal batching to address the issues of inefficient decodes. With chunked-prefills, a prefill request is divided into equal-sized compute chunks. Additionally, decode-maximal batching is utilized by SARATHI to form a batch using one prefill chunk and filling the rest with decodes. This hybrid batch offers units of work that are both compute-saturating and uniform, effectively addressing the problem of inefficient decodes. This method's key insight is that uniformly high computational utilization can be achieved by mixing prefill and decode requests in a single batch. However, because each request comprises only a single prefill phase followed by multiple decode phases (for each generated token), there may not always be enough prefilled requests to create a hybrid batch of prefills and decodes. Chunked-prefills enable the construction of multiple hybrid batches from a single prefill request, thereby increasing the coverage of decodes that can accompany a prefill. In this hybrid batch, the single prefill chunk ensures high GPU utilization, while the decode phase requests 'piggyback' along. Given an average prefill-to-decode token ratio for an LLM application, a prefill chunk size was selected to maximize the overall performance. The hybrid batches generated in SARATHI had uniform computing requirements. Therefore, when utilized with pipeline parallelism, SARATHI ensures well-balanced microbatches, significantly reducing pipeline bubbles. The experiments demonstrate that for LLaMA-13B on A6000, SARATHI enhances the decoding throughput by up to 10×, leading to an up to 1.33× improvement in end-to-end throughput. Similarly, for LLaMA-33B on A100, the decoding throughput improves by 4.25×, resulting in a 1.25× improvement in end-to-end throughput.

\textbf{SPEED} \cite{hooper2024speed} introduces another method to accelerate inference for Transformer decoder models by leveraging speculative execution and parameter sharing. The motivation behind SPEED stems from the inherent slowness of decoding in Transformer models, which is primarily due to their autoregressive nature. While speculative execution promises to improve latency, it typically entails loading different parameters for each token, resulting in overhead. SPEED addresses this challenge by employing cyclic parameter sharing, ensuring that the same parameters are utilized for different tokens in parallel speculative pipelines. Moreover, invalid predictions are corrected before output, enhancing the system's overall accuracy. Experimental results demonstrate that SPEED achieves up to a 3x reduction in latency compared to baseline Transformer models while maintaining accuracy across translation and summarization tasks. Furthermore, SPEED enables the use of deeper models for a fixed model size, thereby enhancing accuracy without increasing latency.

\textbf{TriForce} \cite{sun2024triforce} proposes a scalable and robust speculative decoding system that integrates retrieval-based drafting and hierarchical speculation. LLMs inherently require access to the entire key-value (KV) cache, which stores intermediate key-value states from previous contexts to avoid recomputation. Nevertheless, the phenomenon of sparsity within the attention blocks \cite{liu2023deja} suggests that only a small portion of the KV cache could function as a draft cache. This insight leads to the development of retrieval-based drafting, enhancing acceptance rates during self-speculative decoding. In this method, the KV cache is partitioned into smaller chunks, and during retrieval, attention is computed between a query and the average key cache in each chunk. This process identifies relevant chunks for selective retrieval based on importance scores within a specified resource limit. Unlike passive approaches, this method actively prioritizes critical information, emphasizing relevance over recency.

Another challenge LLMs face is the need to load the whole model weights, which presents an extra bottleneck. To address this challenge, TriForce introduces a hierarchical system. This system utilizes a secondary lightweight model equipped with StreamingLLM cache \cite{xiao2024efficient} to predict initial speculations for the target model, leveraging a retrieval-based draft cache. This draft cache is an alternative to the target model's complete KV cache during drafting. By establishing this sequential speculation hierarchy, the method substantially reduces drafting latency, thereby enhancing the overall inference speed. Experiments show that TriForce not only achieves significant speedups for LLaMA2-7B-128K, reaching up to 2.31× on an A100 GPU, but also exhibits scalability in handling even longer contexts.

    \section{Early Exiting}\label{sec:early}
    
The concept of early exiting was first introduced within the adaptive computation framework across various deep neural network architectures, particularly within encoder-based transformer models \cite{teerapittayanon2017branchynet, graves2017adaptive, liu2020faster, xin2020deebert, zhou2020bert, xin-etal-2021-berxit, zhu-2021-leebert, 10.1145/3534678.3539132,  hu2023smartbert}. This approach is based on observing samples with varying difficulty levels, suggesting that large models may overcalculate simple inputs. In contrast, smaller ones may face challenges with complex samples \cite{liu-etal-2020-fastbert}. Subsequently, \cite{elbayad2020depthadaptive} suggests a decoder architecture that dynamically adjusts the number of layers based on each input to address the high inference latency problem in autoregressive language models. The model's central feature is an internal confidence function that generates a confidence score based on hidden states in intermediate layers. This score is used to decide whether to exit based on predefined early-exiting thresholds. This approach allows the model to generate output faster, even before completing all computations. While effective, this approach poses its own challenges in designing early exiting schemes for language models, where novel techniques are introduced. In the following section, we discuss various methods that tackle these challenges.
 
Following \cite{elbayad2020depthadaptive}, \cite{schuster2022confident} proposes \textbf{Confident Adaptive Language Modeling (CALM)}, a framework aimed at dynamically allocating varying computational resources per input and generation timestep. To outline the problem, consider a Large Language Model with \(L\) layers. Within the early exiting framework, the model initially computes \(p(y_{t+1} | d_{i}^t) = \text{softmax}(W_i d_{i}^t)\) during token \(t\) processing. Then, a confidence score is computed based on a selected confidence measure for intermediate layers \(i < L\), denoted as \(c_{i}^t \in [0, 1]\). If this score exceeds a predefined threshold \(\lambda_{i}^t \in [0, 1]\), the model exits early. This strategy enables the model to avoid full computation across all layers at all times. Consequently, the prediction \(y_{t+1}\) is greedily chosen based on:
\begin{equation}
y_{t+1} :=
\begin{cases}
\arg\max p(y_{t+1} | d_1^t) & \text{if } c_{1}^t \geq \lambda_{1}^t, \\
\arg\max p(y_{t+1} | d_2^t) & \text{if } c_{2}^t \geq \lambda_{2}^t, \\
\vdots & \\
\arg\max p(y_{t+1} | d_L^t) & \text{otherwise}.
\end{cases}
\tag{1}
\end{equation}

The authors explored three different confidence measures:

\begin{itemize}
\item \textbf{Softmax Response}: This method quantifies the difference between the top two values of \(\text{softmax}(W_i d_{i}^t)\), as mentioned previously.

\item \textbf{Hidden-state Saturation}: This approach relies on cosine similarity between \(d_{i}^t\) and \(d_{i-1}^t\) for \(i > 1\). The earliest possible exit occurs at the second layer (\(t\)), unless \(\lambda = 0\). This method aims to detect early saturation events of the hidden state \cite{geva-etal-2022-transformer}.

\item \textbf{Early Exit Classifier}: Here, a dedicated linear model \(M\) is trained to predict the probability of exiting based on the current hidden state \(d_{i}^t\). This method is highly efficient during inference and introduces minimal additional parameters. It employs a per-layer independent cross-entropy loss, and the average is taken across \(L - 1\) layers.
\end{itemize}

Additionally, they proposed a calibration method to define the exit thresholds (\(\lambda_{i}^t\)). This method builds on recent research regarding distribution-free uncertainty quantification, such as \cite{angelopoulos2022gentle}. For this problem, they used the "Learn then test(LTT)" framework \cite{angelopoulos2022learn}, which reframes hyper-parameter selection as a multiple testing problem, along with the monotonic behavior of confidence measures. A sequence of descending thresholds is defined as \(\lambda_1 > \lambda_2 > \ldots > \lambda_k\) with a relatively coarse step size (e.g., increments of 0.05). For each \(\lambda_j\) in order, \(p_j\) is computed, and \(H_j\) is rejected if \(p_j \leq \epsilon\). The first time \(H_j\) is not rejected, the search is terminated, and \(\lambda_{j-1}\) is returned to be used as the calibrated threshold.

\textbf{Fast and Robust Early-Exiting (FREE)} introduced by \cite{bae2023fast} to address the challenges of conventional early-exiting frameworks. These challenges contain overhead computation for confidence scoring and exit thresholds and the risk of losing necessary key-value states for processing subsequent tokens after early exiting. FREE incorporates a shallow-deep module and synchronized parallel decoding, which leverages the advantage of obtaining predictions from different depth models.
A shallow-deep module splits computation paths into two models: a shallow model with a defined number of early layers and a deep model with all layers. During decoding, a synchronized parallel process collects early-exited tokens that only pass through the shallow model until a non-existing token is encountered. In previous works \cite{elbayad2020depthadaptive, schuster2022confident}, a state copying mechanism was employed to address the loss of key-value states for layers after exiting. The synchronized parallel method is an alternative to this mechanism by directly computing multiple hidden states, similar to processing a single token at a time. Additionally, FREE proposed a novel adaptive threshold estimator using parallel decoding outputs to determine the proper confidence threshold for each dataset. By utilizing a Beta mixture model (BMM), this estimator captures the correlation between confidence scores and prediction alignment of the shallow and deep models. Overall, FREE offers a more efficient decoding process than autoregressive token decoding and demonstrates its efficacy in extensive generation tasks.

 Sun et al. suggest another approach called  \textbf{Hash-based Early Exiting (HASH EE)} \cite{sun-etal-2022-simple}. Heuristic metrics such as internal output entropy are typically used to measure instance difficulty, but they suffer from issues like generalization and the need for threshold tunning. In contrast, adopting a "learning to exit" approach or predicting instance difficulty directly is more appealing. Despite some efforts to incorporate such "learn-to-exit" modules, it remains uncertain whether and how effectively instance difficulty can be learned. To address this, HASH EE replaces learn-to-exit modules with hash functions to assign each token to a fixed exiting layer, based on the observation that if a training instance \(x_i\) is predicted to exit at layer \(l\), then an inference instance \(x_j\) that is similar to \(x_i\) should also be predicted to exit at layer \(l\). Unlike previous methods, this removes the need for internal classifiers or extra parameters, making it more efficient. HASH EE offers several advantages: (a) It enables token-level early exiting without supervision, rendering it applicable across various tasks, including language understanding and generation.
 (b) modifying the hash function can easily adjust the speed-up ratio. (c) This method accelerates model inference on a per-batch basis rather than on a per-instance basis.

This study  \cite{kong-etal-2022-accelerating}  proposes a unified framework called \textbf{Multi-perspective early exiting (MPEE)}. Existing approaches mainly focus on the vertical aspect of the architecture, determining which layers should be utilized for inference. In contrast, they overlook the horizontal aspect, neglecting the selection of tokens within each layer for computation. This oversight results in redundancy during adaptive inference. To overcome this, MPEE considers both horizontal and vertical perspectives to accelerate inference. This framework contains layer-wise early exiting (EE) for the vertical perspective and sequential token-wise EE for the horizontal perspective. Specifically, the vertical aspect leverages recycling EE classifier memory and weighted self-distillation to enhance the performance of EE classifiers. Meanwhile, the horizontal perspective utilizes recycling class attention memory to highlight informative tokens, while less informative ones are selectively truncated using weighted fusion, isolating them from subsequent computations. The proposed framework achieves a more balanced tradeoff between performance and efficiency by unifying horizontal and vertical EE. Extensive experimental results demonstrate that MPEE outperforms existing methods in terms of accelerated inference with competitive performance.

\textbf{Predictive Pipelined Decoding (PPD)} \cite{yang2023predictive}  similar to early exit strategies, uses intermediate representations to make predictions about the next token before the entire sequence is processed. However, unlike typical early exit strategies where the processing stops after making a prediction, PPD continues decoding the current token. PPD predicts the distribution of the next token at an intermediate layer and launches parallel sub-processes for the top-k tokens. Meanwhile, the main process continues to the final layer. If a sub-process token matches the token generated by the final layer, PPD continues with that sub-process. Otherwise, it proceeds with the output from the final layer. Theoretical analysis demonstrates that PPD has the potential to achieve 10-37\% lower latency by utilizing 1.5-5x more compute resources. Empirical results show that the token prediction match rates at intermediate layers indicate the potential for acceleration. Furthermore, the match rate increases with more layers, top-k values, and specialized layer classifiers.

\textbf{ConsistentEE}\cite{zeng2024consistentee} approaches early exiting as a reinforcement learning problem. It introduces a policy network to determine if an instance should exit or continue processing. The training goal of ConsistentEE mandates that each instance is accurately predicted by at least one internal classifier. Furthermore, it introduces the idea of a Memorized Layer to measure an instance's difficulty level. This Memorized Layer is integrated into the design of the reward function, allowing "easy" instances to prioritize acceleration while "hard" instances focus more on accuracy.
For the reward function design, ConsistentEE considers both accuracy and acceleration. To ensure accuracy, the loss of the internal classifier is incorporated into the reward function, assigning higher reward values to instances with lower losses. Additionally, ConsistentEE involves the layer depth at which an instance exits into the reward function to account for acceleration. While one can introduce a trade-off coefficient to balance accuracy and acceleration in the reward function, it argues that instances of different hardness levels should assign different weights to accuracy and acceleration.

ConsistentEE observes that "easy" instances generally can be classified correctly at shallow layers, and these instances should exit as early as possible once they are classified correctly. On the other hand, "hard" instances typically can be classified correctly at deeper layers, and they should prioritize accuracy over acceleration at early layers. Hence, ConsistentEE also incorporates the hardness level of an instance into the reward function design. The experimental results show this approach can outperform competitive baselines.

\textbf{SkipDecode} \cite{delcorro2023skipdecode} introduces an EE approach for batch inferencing and KV caching. Existing token-level early exit methods work well for real-time inference tasks, but they struggle when processing batches of data or utilizing Key-Value caching because they have to wait until all tokens in a batch finish processing before stopping. This study addresses the problem by establishing a singular exit point for every token in a batch at each sequence position. A unified exit point is set up for all tokens within a batch at a specific sequence position. Furthermore, a generation policy is designed with monotonically decreasing exit points as the sequence progresses, assuming that subsequent tokens require less computational effort. This strategy also removes the necessity to recalculate Key-Value caches for preceding tokens, substantially reducing computational cost. Another issue with early-exiting methods is that tokens may exit the model early once their hidden states have saturated. Nonetheless, in generative models, this can lead to problems if some tokens exit before others, resulting in a loss of context for later tokens. To mitigate this, they propose skipping instead of early exiting, allocating computation to higher layers to ensure all tokens effectively utilize the available context. SkipDecode achieved 2x to 5x inference speedups in the experiments with negligible regression across various tasks, utilizing OPT models of 1.3 billion and 6.7 billion parameters.

The \textbf{EE-LLM} framework, as described by \cite{chen2024eellm}, introduces an infrastructure for early-exit language models utilizing pipeline parallelism. This system is implemented within Megatron-LM \cite{shoeybi2020megatronlm} to facilitate large-scale training and inference of early-exit LLMs with 3D parallelism. Pipeline parallelism divides a deep model into multiple stages along its depth dimension. Each batch of data is partitioned into smaller parts known as microbatches, which are then processed across these stages. Each stage handles the forward and backward computations for its micro-batch and communicates the results with other stages. This approach presents a challenge for early-exit models, as the training loss typically combines losses from different exits, now located across various pipeline stages. To address this challenge, the authors introduced a lightweight method that guides each pipeline stage to accurately calculate the desired gradients without incurring additional communication overhead between stages. Initially, each stage independently computes gradients as if it operates with its own parameter set. Subsequently, the gradients for tied parameters are summed up and synchronized across stages to ensure consistency.

During the inference phase of early-exiting models, an additional challenge arises in autoregressive generation tasks, where tokens are produced sequentially based on previously generated tokens using the attention mechanism. Early-exit inference contradicts KV caching, a method that stores the keys and values of previously generated tokens at each layer. However, if the current token is generated through early exiting at a particular layer, its KV caches in subsequent layers are absent. To address this issue, this paper devised two approaches. One technique involves KV recomputation, wherein the forward pass is executed with a list of recent tokens during token generation. The other method is based on a new type of pipeline parallelism, which concurrently processes the forward pass of the current token at a specific pipeline stage alongside some KV-related computation (if applicable) of previous tokens at later stages. 

\textbf{LayerSkip} \cite{elhoushi2024layer} combines exiting early with speculative decoding to propose a self-speculative decoding approach that does not require an additional model or auxiliary layers. This approach has three different stages:
1. Training using Layer Dropout \& Early Exit Loss
2. Inference using Early Exit
3. Verification and Correction using Speculative Decoding.
Typically, deep learning models don't try to predict their final output early; instead, they distribute their computational load evenly across all layers \cite{voita2023neurons}. In this manner, early exiting tokens suffer from a lack of accuracy. To tackle this issue, it is suggested that layer skipping (layer dropout) be implemented during training to enable earlier predictions. To be more exact, higher dropout rates are designated for later layers, whereas lower dropout rates are applied to earlier layers. Tokens can predict and exit early in this fashion. 

Additionally, the authors explore strategies to address the slowdown in training and reduction in accuracy caused by adding early exit loss to all layers during training. To overcome this, they proposed two curricula. The first is a rotational early exit curriculum, where early exit is enabled at every $R$ layer, rotating at each iteration. The second is a gradual early exit curriculum, gradually introducing early exit loss from the last layer to the first layer, one layer at a time, every $\frac{T}{2L}$ iterations (layer $L$ and training iteration $T$). The early exit tokens are then verified via a self-speculative decoding process that consists of two key steps: (1) Self-Drafting, which involves using the early exit to draft tokens from the same model, and (2) Self-Verification, which uses the remaining layers to validate the prediction. A unique mechanism called Cache Reuse is developed in order to facilitate reuse in (1) and (2) that unifies the KV cache and stores the exit query. The experiments showed that combining layer dropout and early exit loss with curriculum improved the accuracy of early exit during inference and developed a self-speculative decoding solution resulting in a speedup of up to 1.86×.

    \section{Non-autoregressive Models}\label{sec:non-auto}
    In response to the limitations of sequential execution, a distinct class of strategies focuses on Non-autoregressive (NAR) models. In NAR models, all target tokens are generated simultaneously or in parallel, unlike autoregressive (AR) models, where tokens are generated one at a time in a sequential manner, which accelerates the inference speed. However, because all tokens are generated simultaneously, there's no sequential process to guide decoding termination. Each token is generated independently of the others, without direct knowledge of the other token's position or identity. Therefore, there's no special token or target information to signal the end of the decoding process. This lack of sequential guidance for termination can pose challenges in non-autoregressive models, as they need to find alternative methods to determine when the generation process should stop. This section investigates these models and how they tackle this problem. As a disclaimer, it's important to note that most of the papers discussed in the following section primarily focus on NAR-based methods for machine translation tasks. This focus is inherent to the traditional emphasis of this research domain.

\textbf{The Non-autoregressive transformer model (NAT)} \cite{gu2018nonautoregressive} is recognized as one of the pioneering works in this field. NAT is specifically introduced for machine translation tasks to avoid the autoregressive property and produce outputs in parallel. As mentioned above, while this capability reduces latency during inference, it poses a significant challenge to accuracy because it relies only on source-side information in the decoder and lacks target tokens. Also, it is crucial for models to determine the target sentence's length before decoding generates all words simultaneously. During training and inference, accessing time-shifted target outputs or previously predicted outputs as inputs to the first decoder layer in a parallel manner is impossible. Simply removing inputs or relying only on positional embeddings resulted in poor performance. Therefore, the decoding process is initiated by using copied source inputs from the encoder side. Because source and target sentences often differ in length, NAT proposed a fertility predictor \cite{brown-etal-1993-mathematics} to determine how many times the source token should be copied when constructing the decoder input, with the sum of fertility numbers defining the target sentence length.

At inference time, the model seeks the outputs with the highest probability by analyzing all fertility sequences. However, due to the impracticality of searching the entire fertility space, this paper proposes three heuristic decoding strategies: Argmax decoding, which selects the highest-probability fertility for each input word; Average decoding, which estimates each fertility as the expectation of its corresponding softmax distribution, and Noisy Parallel Decoding (NPD) \cite{cho2016noisy}, which draws samples from the fertility space and computes the best translation for each fertility sequence. Additionally, the use of sequence-level knowledge distillation and policy gradient fine-tuning enhances NAT's training process. Experimental evaluations conducted on WMT English-German/Romanian and IWSLT German-English datasets demonstrate that NAT, particularly with NPD, achieves quality comparable to autoregressive transformers while offering over a tenfold reduction in inference latency. Furthermore, analysis indicates that NPD enhances diversity and mitigates issues such as repeated words, positively impacting performance and training stability.

Following that idea, \cite{guo2018nonautoregressive} proposed two different methods to enhance the decoder input for NAT models. These methods enable tokens to be fed with some target information into the decoder, thereby aiding the model in learning the training data more effectively. The authors realized that while the encoder of AR models and that of NAT models are the same, the differences lie in the decoder, which lacks target-side information. Therefore, the basic idea is to feed target-side tokens directly as the decoder inputs. They introduced two approaches to generate the decoder input sequence with coarse target-side information. The first method relies on a pre-trained phrase table, a resource commonly used in machine translation tasks. Using the phrase table, this approach directly translates each source token into target-side tokens. Essentially, the phrase table serves as a lookup dictionary, providing explicit mappings from source tokens to their corresponding target-side counterparts. The second method, known as embedding mapping, operates by linearly transforming the embeddings of source tokens into the target-side embedding space. This mapping is learned end-to-end during training to minimize the L2 distance between the mapped source and target embeddings at the sentence level. Additionally, an adversary loss is employed at the word level to ensure that the mapped source embeddings retain relevant information for the target side while minimizing any introduced distortion. Overall, both methods aim to enrich the decoder input with target-side information, thereby enhancing the training process and improving the performance of non-autoregressive models.

Using latent variables within the model is an alternative approach to reducing the dependency on the target side in NAR models, \cite{kaiser2018fast} introduced the concept of a Latent Transformer (LT) incorporating discrete latent variables, which enables more parallelizable decoding. Traditional decoding methods often suffer from slow speed as they depend on previous tokens. However, introducing shorter discrete latent sequences that encapsulate the essential information from the sequence makes more efficient target generation possible. The target sequence is initially encoded into a shorter discrete latent sequence, which the encoder then predicts autoregressively. Subsequently, the target is decoded in parallel from both the latent sequence and the encoder. Nonetheless, utilizing discrete latent variables can pose challenges during model training. To address this, the authors proposed a new discretization technique known as Decomposed Vector Quantization (DVQ). DVQ decomposes the encoder output into smaller slices, allowing for more efficient training and utilization of embedding vectors, which is particularly beneficial when the size of the discrete latent space is large. Experimental results demonstrate that the LT model with DVQ outperforms non-autoregressive baselines in machine translation tasks, achieving up to a 10x decoding speedup.

In \cite{ma-etal-2019-flowseq}, the authors also present \textbf{FlowSeq}, another model for non-autoregressive sequence generation that utilizes latent variables. Non-autoregressive models try to decouple the dependencies of decoding history during generation, which can result in poor performance. FlowSeq, however, use generative flow \cite{rezende2015variational}, a method capable of transforming simple distributions like Gaussians into complex ones through invertible transformations. By encoding the prior distribution \( p_\theta(z|x) \) using generative flow, FlowSeq introduces meaningful latent variables \( z \) into the non-autoregressive generation process, thereby modeling dependencies between output tokens while enabling efficient parallel decoding. This approach addresses the limitations of naively assuming token independence in basic non-autoregressive models and represents a significant advancement in the field of sequence-to-sequence modeling. FlowSeq performs similarly to the best non-autoregressive models, with decoding time remaining nearly constant regardless of sequence length.

Building upon the idea of integrating non-autoregressive decoding with latent variables, \cite{lee-etal-2018-deterministic} propose a novel non-autoregressive neural sequence model employing iterative refinement. This approach applies to various sequence generation tasks beyond machine translation. The model operates as both a latent variable model and a conditional denoising autoencoder, integrating a learning algorithm that combines lower-bound maximization and reconstruction error minimization. To address the trade-off between generation latency and quality, the authors devise an iterative inference strategy with an adaptive number of steps. They further specialize the learning algorithm to interpret the entire model as a latent variable model, with each refinement step viewed as denoising. This methodology, implemented using the Transformer architecture, is evaluated on both machine translation and image caption generation tasks. Results demonstrate that the proposed non-autoregressive model performs similarly to the autoregressive counterpart while significantly accelerating decoding. Moreover, the qualitative analysis highlights the efficacy of iterative refinement in gradually enhancing target sequences across multiple steps.

In \cite{akoury-etal-2019-syntactically}, authors introduce the \textbf{Syntactically Supervised Transformer (SynST)}. This model automatically predicts a chunked parse tree before generating the entire target sentence conditioned on the predicted parse. The key idea is that they use syntactic information as a proxy to the learned discrete latent space of the LT. The model uses syntax as a framework for generating the target sentence. To train the model for syntactic prediction, an external parser in the target language is utilized to provide supervision. Instead of predicting the entire linearized parse minus the terminals, which would significantly increase the number of autoregressive steps and slow down decoding, a chunking algorithm is employed on the constituency parse. This algorithm involves selecting a maximum chunk size and traversing the parse tree in an in-order manner. At each node, if the number of leaves spanned by that node is within the maximum chunk size, a descriptive chunk identifier is appended to the parse sequence. Each unique chunk identifier, formed by the concatenation of the constituent type and subtree size, is considered an element of the first decoder's vocabulary. This approach balances syntactic expressivity with decoding efficiency by limiting the vocabulary size based on the maximum chunk size and constituent types. Controlled experiments demonstrate that SynST surpasses competing non-autoregressive methods in both BLEU scores and wall-clock speedup.

Another research direction \cite{guo2020fine, zhan2022non} has focused on fine-tuning NAR models with autoregressive (AR) models. Considering the higher accuracy and ease of training offered by AR models compared to NAR models and their similar configurations, it's logical to enhance the accuracy of NAR models by transferring knowledge from well-trained AR models through fine-tuning. In their work, \cite{guo2020fine} introduced the idea of curriculum learning \cite{10.1145/1553374.1553380} into NAR model training. This involved gradually transferring the decoder input from AR to NAT, facilitating a smooth transition between the two training strategies. They regarded AR training as the easier task and NAR training as the harder task. A curriculum was designed to gradually transfer the decoder input and attention mask from an AR decoder to a NAR decoder. In this work, only forward dependency in AR models was used to initialize model parameters for NAR.

Zhan et al. extended this idea by proposing a novel method called \textbf{Dependency-Aware Decoder (DePA)} to refine target dependency modeling \cite{zhan2022non}. An effective forward-backward dependency modeling approach (FBD) was introduced as an autoregressive forward-backward pre-training phase before NAR training. They designed this method from two aspects: decoder self-attention and decoder input. Firstly, they propose a pre-training phase where the NAR decoder learns bidirectional target dependencies gradually. Secondly, they implement a novel attentive transformation process to convert the decoder input from the source language representation space to the target language representation space. This transformation enables the decoder to model target dependencies more effectively.

Instead of fine-tuning a NAR model, \cite{su-etal-2021-non} show that BERT \cite{devlin2018bert}, a pre-trained language model, can serve as the foundational framework for a Non-Autoregressive model. They also introduce strategies to address two common issues in traditional NAR models: target length prediction and the conditional independence of target tokens. Their proposed decoding mechanism simplifies the process by enabling the model to determine the output length dynamically. Instead of requiring explicit predictions of the output length, the model emits a [EOS] token at any position to signal the end of the generated sequence. This eliminates the need for additional steps such as output length prediction or post-generation re-ranking. Additionally, many existing models in the non-autoregressive generation domain assume that token predictions at different positions are independent. However, this assumption often results in the generation of sequences with repetitive and ungrammatical patterns. To mitigate this, the authors propose a context-aware learning objective, encouraging the model to generate diverse tokens at adjacent positions, thereby reducing the likelihood of generating repetitive sequences and enhancing overall grammaticality. Furthermore, to leverage the speed advantage of the proposed model, they introduce a novel decoding strategy called "ratio-first," which is particularly beneficial when output lengths can be roughly estimated in advance. This strategy utilizes only a portion of the hidden information from the source, determined by the source length and a predetermined ratio, enhancing inference speed without sacrificing quality.

The authors of \cite{wang2018semiautoregressive} introduce the \textbf{Semi-autoregressive transformer (SAT)}, which preserves the autoregressive property at a global level while relaxing it locally to enable parallel sequence generation. The SAT architecture closely resembles that of the transformer, but with some modifications in the decoder. Unlike the Transformer, the SAT utilizes a group-level chain rule to avoid dependencies between consecutive words within the same group. In SAT, the prediction of \( y_t \) is based on \(y\_{t-K}\), referred to as long-distance prediction. In contrast to the strict causal mask used in the Transformer decoder, the SAT employs a relaxed causal mask. This coarse-grained lower triangular matrix allows the model to access information from later steps within the same group, thus improving prediction accuracy. By adopting the group-level chain rule, long-distance prediction, and relaxed causal mask, they extended the Transformer to the SAT model. Experimental results on English-German and Chinese-English translation demonstrate that, compared with non-autoregressive methods, the SAT achieves a better balance between translation quality and decoding speed.

\textbf{Fast structured decoding for sequence models} \cite{sun2019fast} introduces a new framework to bridge the gap between non-autoregressive and autoregressive sequence models. The key idea is to incorporate a structured inference module into non-autoregressive decoders, which directly models the distribution of target sequences. Specifically, the paper introduces the use of linear-chain Conditional Random Fields (CRF) \cite{10.5555/645530.655813} to model richer structural dependencies between adjacent words in the target sequence. By doing so, the CRF-based structured inference module improves decoding consistency on the target side. Traditional sequence models typically have large vocabulary sizes, which pose challenges for traditional CRFs. To address this issue, the paper presents two effective approximation methods for CRF: low-rank approximation and beam approximation. Low-rank approximation reduces computational complexity by representing CRF parameters in lower-dimensional factors, while beam approximation limits the search space to the most promising label sequences, mitigating computational costs. Additionally, the proposed framework includes a dynamic transition technique to leverage contextual information from the hidden states of the non-autoregressive decoder, thereby enhancing the expressive power of the structured inference module. Overall, this framework aims to enhance the performance of non-autoregressive sequence models in tasks such as machine translation by incorporating structured inference methods that model dependencies between adjacent words and leverage contextual information from the decoder.

In \textbf{Mask-Predict} \cite{ghazvininejad2019maskpredict}, authors propose an innovative solution to the primary challenges faced by non-autoregressive models. Initially, they introduce conditional masked language models (\textbf{CMLMs}), which are encoder-decoder architectures trained with a masked language model objective \cite{devlin2018bert}. This adaptation enables the model to concurrently predict any subset of masked words in the target. Additionally, they present a novel decoding algorithm that leverages the order-agnostic nature of CMLMs to support highly parallel decoding. The mask-predict decoding algorithm operates by decoding an entire sequence in parallel within a constant number of cycles. During each iteration, the algorithm identifies a subset of tokens to mask based on the model's confidence level. These masked tokens are then predicted simultaneously using an underlying conditional masked language model (CMLM). By masking tokens where the model is uncertain and utilizing previous high-confidence predictions, the model can re-predict challenging cases with additional information. This iterative process allows the model to reconsider word choices within a rich bi-directional context, ultimately leading to high-quality translations. Moreover, the algorithm's capacity to make significant parallel changes at each step enables it to converge on a high-quality output sequence in a sub-linear number of decoding iterations. Experiments further demonstrate the effectiveness of CMLMs with Mask-Predict over previous non-autoregressive models, particularly in machine translation tasks, achieving approximately three times faster decoding while approaching autoregressive model performance.

Another line of research addresses the issue through the lens of \textbf{Jacobi decoding} algorithms. In their study, \cite{santilli-etal-2023-accelerating} introduces an innovative technique to speed up autoregressive decoding in machine translation tasks while maintaining translation quality. They propose a parallel formulation of the conventional greedy autoregressive decoding process, leveraging Jacobi and Gauss-Seidel fixed-point iteration methods \cite{ortega1970iterative} for rapid inference.

The greedy decoding procedure for all tokens can be expressed as follows:
\begin{equation}
\begin{cases}
    y_1 = \arg\max p_{\theta}(y_1 | x) \\
    y_2 = \arg\max p_{\theta}(y_2 | y_1, x) \\
    \vdots \\
    y_m = \arg\max p_{\theta}(y_m | y_1:y_{m-1}, x)
\end{cases}
\tag{2}
\label{eq:greedy_decoding}
\end{equation}

By defining \( f(y_i, y_1:i-1, x) = y_i - \arg\max p_{\theta}(y_i | y_1:i-1, x) \), the system of Equations \ref{eq:greedy_decoding} can be rewritten as:
\begin{equation}
\begin{cases}
    f(y_1, x) = 0 \\
    f(y_2, y_1, x) = 0 \\
    \vdots \\
    f(y_m, y_1:m-1, x) = 0
\end{cases}
\tag{3}
\end{equation}

This system comprises \( m \) non-linear equations (each utilizing a neural network) with \( m \) variables. To solve them, this approach randomly guesses the next n tokens in a sequence (n-token sequence hereinafter) from an input prompt. The n-token sequence and prompt are then fed to the LLM to update itself iteratively. Eventually, the n-token sequence converges to the same output generated by AR decoding under a greedy strategy. The evolution of the n-token sequence forms a Jacobi trajectory between a randomly initialized sequence and the n-token sequence generated by AR decoding (i.e., the fixed point). The paper also suggests that this approach can be further developed by drawing from the literature on numerical methods for solving non-linear equations. With appropriate stopping conditions, quality guarantees over the output can be achieved. The paper presents three algorithms (Parallel Jacobi (PJ) Decoding, Parallel GS-Jacobi (PGJ) Decoding, and Hybrid GS-Jacobi (HGJ) Decoding) that leverage these fixed-point iteration methods to accelerate decoding in machine translation tasks. The utilization of hyperparameters enables the control of block size and early stopping. Evaluation across diverse languages and models illustrates that parallelization can enhance speed by up to 38\% compared to the standard autoregressive decoding method and nearly double the speed when exploiting parallel computing resources. Also, when compared to a selection of NAT models, HGJ demonstrates the most efficient use of computational resources, achieving the best cost-benefit ratio even when considering additional inference costs. Moreover, while NATs typically result in degraded translation quality compared to their autoregressive baseline, the method ensures the same high-quality autoregressive decoding, surpassing standard NAT models.

As observed in practical implementations, Vanilla Jacobi decoding for LLMs exhibits only slight speed improvements compared to autoregressive decoding, typically achieving an average speedup of 1.05×, as demonstrated by \cite{santilli-etal-2023-accelerating}. This marginal enhancement occurs because LLMs often struggle to produce accurate tokens if there are errors in their preceding tokens. \textbf{Consistency Large Language Models (CLLMs)} \cite{kou2024cllms}
address this by proposing an approach to achieve rapid convergence from any state to the fixed point on a Jacobi trajectory. This method focuses on refining the LLMs to generate multiple subsequent tokens of a prefix simultaneously, aiming to match the output of autoregressive decoding in just one step. However, the preliminary experiments revealed that learning this task in a single step is challenging, particularly with large token sequences. They incorporate intermediate points along the Jacobi trajectory that contain more accurate tokens to tackle this issue. Specifically, the learning process mirrors AR modeling for the second-to-last point on the trajectory, where the target LLM performs well even without adaptation. This approach achieves significant speedup with minimal performance degradation and does not require additional memory costs for auxiliary model components. The learning strategy involves training the model with two loss terms: a consistency loss, where the model is trained to map any point on the Jacobi trajectory to the fixed point, and an AR loss to maintain the distribution of the target LLM and ensure generation quality. This method has exhibited substantial improvements in generation speed, ranging from 2.4 to 3.4 times faster, while preserving high-quality results across various benchmark tests, both in domain-specific and open-domain contexts.

\textbf{Skeleton-of-Thought (SoT)} \cite{ning2023skeletonofthought} is another approach aimed at accelerating inference in LLMs through parallel decoding. Traditional autoregressive decoding methods are often slow, whereas certain LLMs can score output tokens in parallel. SoT capitalizes on this parallel scoring capability by initially having the LLM generate a concise skeleton of the answer, followed by simultaneous expansion on each point. It consists of two stages: the skeleton stage, where the LLM produces concise skeleton points, and the point-expanding stage, where the LLM simultaneously expands on each point as a batch. For API models, parallel API calls are executed, while point-expanding requests are decoded together for open-source models. Experimental results show that SoT reduces inference time by up to 2.4x without compromising quality. Moreover, analysis suggests that SoT enhances the diversity and relevance of generated answers. Additionally, the paper proposes further optimization with a router to selectively trigger SoT, introducing SoT-Router, which maintains efficiency gains while enhancing answer quality. Overall, SoT offers a promising approach to accelerating LLM inference, providing significant efficiency gains and improved answer quality.

    \section{Discussion and Limitations}\label{sec:discussion}
        This study comprehensively examines various accelerated generation techniques utilized in large language models (LLMs), emphasizing their potential to expedite the generation process without compromising the accuracy and quality of generated outputs. Among the various techniques discussed, speculative decoding stands out due to its promise of enhancing efficiency through parallel prediction and verification processes, potentially circumventing the inherent sequential bottlenecks of traditional approaches. Our survey categorizes speculative decoding into distinct phases—drafting and verification—each with unique challenges and tailored solutions. Another approach is Early exiting strategies that speed up inference by stopping generation processes once a sufficiently confident prediction is made, leveraging the observation that tokens differ in complexity. By detecting simpler tokens and exiting computations early, these strategies can make LLMs more adaptable to real-time applications. Furthermore, non-autoregressive decoding techniques use parallel processing to accelerate inference, enabling rapid generation of responses. By relaxing the rigid left-to-right generation order, these methods facilitate more efficient parallelization and address a longstanding challenge in sequence-to-sequence modeling.
        
    Despite the advancements, these accelerated generation approach introduces several challenges and limitations. One concern is that the computational overhead of implementing advanced generation techniques is a significant barrier, especially in resource-constrained environments or applications requiring real-time processing. While speculative decoding aims to reduce overall decoding time, the increased computational demand for parallel processing can offset these gains, particularly when the speculative paths diverge significantly from the greedy path, leading to wasted computational resources. Moreover, specific speculative decoding methods require training a separate draft model to generate multiple candidate outputs, thereby increasing the computational burden. Similarly, some early exiting methods require an extra model head to predict token difficulty or compute a confidence function for determining when to terminate generation, further complicating computational complexity. In non-autoregressive decoding, fine-tuning the model to predict tokens in parallel may compromise accuracy compared to autoregressive approaches.
    
    Moreover, the complexity of integrating new algorithms with existing LLM architectures without extensive modifications remains a substantial hurdle. The need for additional models or modifications to the generation mechanism can complicate the implementation and scalability of these methods in standard LLM frameworks. This aspect requires further research into integration techniques that can leverage accelerated generation benefits without disruptive changes to the model architecture. 
    
    Finally, the field of minimizing computational overhead due to generation methods remains ripe for exploration. Future research could focus on optimizing the efficiency of these methods, perhaps through more intelligent speculative path selection or better integration with lightweight, adaptive model architectures. While recent advancements in these methods have improved the capabilities of LLMs to unprecedented levels, addressing these limitations is crucial for realizing their full potential.
    
    \section{Conclusion}\label{sec:conclusion}

    This survey paper explores accelerated generation algorithms that aim to reduce the latency associated with LLM inference. We explore various strategies that enable parallel or more efficient token generation to speed up the decoding process while maintaining the quality of generated outputs. Our investigation includes examining different techniques, such as speculative decoding, early exiting strategies, and non-autoregressive approaches, to identify methods that can improve the speed and efficiency of LLM inference. We have identified key trends and advancements in each category through a detailed literature analysis, providing valuable insights into existing approaches' strengths and limitations. Figure \ref{fig:taxonomy} presents a comprehensive taxonomy that researchers can use as a roadmap to navigate the array of LLMs acceleration methodologies. Our review emphasizes the importance of addressing latency issues in LLMs and highlights the potential of novel strategies and algorithms to enhance their efficiency and applicability in various domains. Better accelerated generation techniques can lead to more effective and scalable LLMs that meet real-world demands.
  
    \bibliography{main}
    \bibliographystyle{unsrtnat}
	
\end{document}